\documentclass[11pt,a4paper]{article}
\usepackage[hyperref]{acl2018}
\usepackage{times}
\usepackage{latexsym}
\usepackage{rotating}
\usepackage{multirow}
\usepackage{url}

\aclfinalcopy % Uncomment this line for the final submission
\title{ABI Neural Ensemble Model for Gender Prediction\\
{\large Adapt Bar-Ilan Submission for the CLIN29 Shared Task on Gender Prediction}}

{\small \author{Eva Vanmassenhove$^\alpha$,~~
  Amit Moryossef$^\beta$,~~
  Alberto Poncelas$^\alpha$,~~
  Andy Way$^\alpha$,~~
  Dimitar Shterionov$^\alpha$\\\\
    $^\alpha$ ADAPT, School of Computing,
  Dublin City University, Dublin, Ireland\\
  {\tt firstname.lastname@adaptcentre.ie}\\
  $^\beta$ Computer Science Department, 
  Bar-Ilan University, Ramat-Gan, Israel\\
  {\tt firstnamelastname@gmail.com}
}}

\begin{document}
\maketitle
\begin{abstract}

We present our system for the CLIN29 shared task on cross-genre gender detection for Dutch. We experimented with a multitude of neural models (CNN, RNN, LSTM, etc.), more ``traditional'' models (SVM, RF, LogReg, etc.), different feature sets as well as data pre-processing. The final results suggested that using tokenized, non-lowercased data works best for most of the neural models, while a combination of word clusters, character trigrams and word lists showed to be most beneficial for the majority of the more ``traditional'' (that is, non-neural) models, beating features used in previous tasks such as $n$-grams, character $n$-grams, part-of-speech tags and combinations thereof. In contradiction with the results described in previous comparable shared tasks, our neural models performed better than our best traditional approaches with our best feature set-up. Our final model consisted of a weighted ensemble model combining the top 25 models. Our final model won both the in-domain gender prediction task and the cross-genre challenge, achieving an average accuracy of 64.93\% on the in-domain gender prediction task, and 56.26\% on cross-genre gender prediction.

\end{abstract}

\section{Introduction}

In recent years, author profiling (AP) has gained a lot of interest. AP can be described as the task of predicting or identifying demographics (such as gender and age) of an author based on their writing. It has applications of growing importance in different fields such as security, marketing, etc. Of all the demographics, there has been a particular interest in gender profiling which has been part of the shared tasks organized by PAN for the past six consecutive years (2013--2018). Tasks differed from in-domain gender prediction achieving relatively high scores to multi-modal gender prediction and cross-genre gender prediction, initially focusing on English but gradually including other languages such as Dutch, Spanish, and others.

In this paper, we describe our best system submitted for the CLIN29 2019 shared task on gender prediction for Dutch with a particular focus on out-of-domain data, along with all other systems we experimented with. The plethora of trained models and empirical evaluations we conducted for this task is one of our main contributions; the other is the ensemble system which wins both the in-domain and out-of-domain subtasks.

The rest of the paper is structured as follows: Section~\ref{sec:relwork} describes the related work, Section~\ref{sec:experiments} the experimental set-up. Results are discussed in Section~\ref{sec:results}. Finally, the conclusions and ideas for future work are presented in Section~\ref{sec:conclusions}.

%A crucial aspect of this task is designing settings, as they are key to shed light on the core question: are there indicative traits across genres that can be leveraged to model gender in a rather genre-independent way?

\section{Related Work}~\label{sec:relwork}
Over the last years, there has been a significant body of research dedicated to AP, often including gender prediction. The main focus, however, has been on predicting gender using in-domain data for training and testing. The various approaches presented in the PAN shared tasks used preprocessing, lowercasing, stop-word filtering combined with specific linguistic features including character, word and POS $n$-grams, punctuation features, topic modeling and features specific to the given domain (e.g. hashtags or links for Twitter data). The yearly PAN evaluation campaigns (pan.webis.de) have led to the development of state-of-the-art (SOTA) in-domain gender prediction models on Twitter data for English achieving accuracies in between $80\%-85\%$ \cite{alvarez2015inaoe,rangel2015overview,basile2017,rangel2017overview}. 

The PAN 2016 differed from previous gender prediction tasks as it was the first shared task focusing on cross-genre gender prediction. Twitter data was provided for training while the test data was another `unknown' type of social media text. It should be noted, however, that although the test data differed from the training data, all the data still belonged to the broader 'social media' domain. The best scores recorded for gender prediction were $62\%$, $73\%$ and $76\%$ for Dutch, Spanish and English respectively~\cite{rangel2016overview}. An additional analysis of the cross-genre results by Medvedeva et al.~(\citeyear{medvedeva2017analysis}) revealed that the portability of the cross-genre models is only successful when the subdomains are close enough. The PAN-RUS Profiling at FIRE’17 focused on predicting gender across different domains (Twitter, Facebook, essays and reviews) obtaining accuracies between $65\%-93\%$~\cite{litvinova2017overview} depending on the domain. Similarly, in order to capture more domain-independent and thus deeper gender-specific features, the EVALITA 2018 Campaign~\cite{Caselli2018} organized a cross-genre prediction task across five domains (Children Writings, Twitter, YouTube, News, and Personal Diaries) with accuracies ranging between 51\% (YouTube) and 64\% (Children Writings).

An important difference between the two previous tasks on real cross-genre gender prediction and this year's CLIN shared tasks is that, unlike Russian and Italian, gender agreement with the first person is very rare in Dutch\footnote{Exceptions would be certain sentences where the noun agrees in gender with the subject, e.g., `Ik ben een acteur' (masc.) vs `Ik ben een actrice' (fem) [English: `I am an actor/actress']}. In Russian and Italian, verbs, adjectives and nouns (can) reflect the gender of the speaker, which facilitates gender prediction.

% \dimitar{Consecutively, we focused on identifying stylistic and  }

\section{Experimental Setup}~\label{sec:experiments}
In this section, we describe in more detail: the datasets provided~(\ref{subsec:data}) and the different models and features used~(\ref{subsec:models}). 

\subsection{Datasets}\label{subsec:data}

The datasets initially provided belonged to three different domains: Twitter (TW), YouTube (YT) and News (N). For the 3 in-domain scenarios we used $90\%$ of the data for training and $10\%$ for validation. For the 3 out-of-domain scenarios we used all the out-of-domain data available. For example, for the out-of-domain YouTube prediction, we used all the given News and Twitter data and validated on the YouTube data provided. For some of our models (i.e., our second set of submissions), we also used external CSI~\cite{verhoeven2014clips} and TwiSty~\cite{verhoeven2016twisty} data. We made sure there was no overlap between the data provided and the external data we added. An overview of the different scenarios and the amount of training/validation examples can be found in Table~\ref{tbl:dataCLIN}.

\begin{table}[h!]
    \centering
    \footnotesize\begin{tabular}{c|l|r|r|}
    \cline{3-4}
    \multicolumn{1}{c}{\multirow{4}{*}{\rotatebox[origin=c]{90}{\scriptsize IN-DOMAIN}}} & \multicolumn{1}{c|}{}   &  \bf{\# Train} &   \bf{\# Valid}   \\ \cline{2-4}
    & \bf{N  (90-10)}         &  $1~648$	    &	$184$	            \\ \cline{2-4}
    & \bf{TW (90-10)}         &	$18~000$	    &	$2~000$	        \\ \cline{2-4}
    & \bf{YT (90-10)}         &	$13~269$	    &	$1~475$	        \\ \cline{2-4}
    \end{tabular}\newline\newline
    
    \begin{tabular}{c|l|r|r|}
    \cline{3-4}
    \multicolumn{1}{c}{\multirow{3}{*}{\rotatebox{90}{\scriptsize OUT-DOMAIN}}} & \multicolumn{1}{c|}{}   &  \bf{\# Train} &   \bf{\# Valid}   \\ \cline{2-4}
    & \bf{N+TW \textbar YT}   &	$21~832$	    &	$14~744$	        \\ \cline{2-4}
    & \bf{N+YT \textbar TW}	&	$16~576$	    &	$20~000$	        \\ \cline{2-4}
    & \bf{TW+YT \textbar N}	&   $34~744$	    &	$1~832$	        \\ \cline{2-4}
    \end{tabular}\newline\newline

    \begin{tabular}{c|l|r|}
    \cline{3-3}
    \multicolumn{1}{c}{\multirow{2}{*}{\rotatebox{90}{\scriptsize EXTERNAL}}} & \multicolumn{1}{c|}{}   &  \bf{\# Train}\\\cline{2-3}
    & \bf{CSI}   &	$3~113$ \\ \cline{2-3}
    & \bf{TWisty}	&	$68~907$ \\ \cline{2-3}
    \end{tabular}\newline
    
\caption{Number of training and validation examples per scenario.}\label{tbl:dataCLIN}
\end{table}

\subsection{Systems Descriptions}~\label{subsec:models}
Our winning approach consisted of an ensemble model of our strongest models. First, we will first describe all the neural and traditional models we experimented with as well as the effects of different feature sets. After, the ensembling of the best models will be described in more detail.

\par
\textbf{Neural Models}
We experimented with the following neural networks: SpaCy \textit{TextCategorizer} models (SpaCy)~\cite{spacy2}, Convolutional Neural Network (CNN), Long Short-Term Memory (LSTM), Long Short-Term Memory with Attention (LSTMa), Region-based Convolutional Neural Networks (RCNN), Recurrent Neural Network (RNN) and Self Attention (SA). All models were trained with and without frozen \verb|fasttext| embeddings~\footnote{As the fasttext embeddings are extracted from Wikipedia, yet a different domain, the usage of these embeddings does not interfere with the objective of the cross-genre prediction task.}~\cite{joulin2016bag} using a publicly available pytorch implementation.\footnote{\url{https://github.com/prakashpandey9/Text-Classification-Pytorch}}

All neural network models we trained using the following parameters: \textbf{batch size} $= 32$, \textit{hidden state size} $= 256$ and \textit{embedding length} $= 300$; The \textit{learning optimizer} was Adam~\cite{Kingma2014} and we used a \textit{learning rate} $= 2e-5$. For the SpaCy models, the default parameters were used.\footnote{See: \url{https://spacy.io/usage/training} for details.}

\par
\textbf{Traditional Models}
From previous shared tasks it resulted that more traditional statistical models, in general, still outperform neural models. The traditional approaches we experimented with include: Statistical Language Models (LMs), Support Vector Machines (SVMs), K-Nearest Neighbour Classifier (KNN), Logistic Regression Classifier (LogReg), Random Forest Classifier (RF) and Bernouilli Naive Bayes (NB) using the \verb|scikit-learn| toolkit~\cite{scikit-learn}.

We ought to note some differences between our LM-based models and the other approaches. In particular, for the LM-based models, we: (i) split the data into two subsets -- a female and a male subset -- which we (ii) use separately to train two KenLM~\cite{Heafield2013} models, one for each of the two subsets. At training time we build $3$-, $4$-, $5$- and $6$- gram KenLM models with pruning of singleton $2$-grams and above; each paragraph was also segmented into single sentences. For each sentence in a test paragraph, we generated two scores -- one for the female and one for the male models. The scores for a paragraph are the averages over all sentence scores in that paragraph. To classify the test paragraph we identify whether the female or the male KenLM model leads to higher score. This approach contrasts with the rest, where female and male data is used jointly to train a model that predicts (probabilistically) one of two labels.

\par
\textbf{Features} As previous shared tasks showed the usefulness of features such as word $n$-grams, character $n$-grams, Part-Of-Speech (POS) tags, and article counts, we compared how the 'traditional' classifiers performed with such features. We experimented with 4 additional features:
\par
\textit{clusters:} Inspired by the cross-genre task and the necessity for generalization we added a cluster feature. To do so, we clustered words together based on their \verb|fasttext| embeddings and subsequently replaced unique words by their cluster number. Clusters containing only singletons or more than 500 words were removed.

The three following features were inspired by the findings of \citet{keune2012explaining} whose thesis investigated the difference between male and female speech.
\par
\textit{words used more by men:} \citet{keune2012explaining} concluded from her research on the Corpus Gesproken Nederlands (CGN)\footnote{`Corpus of spoken Dutch'} that there were certain words linked to a specific gender. Similar observations were made in \citet{Vanmassenhove2018}, showing that certain words are more frequent in male/female speech. Words such as: `feitelijk', `voornamelijk', `degelijk', `oorspronkelijk', `tamelijk', `onmiddellijk', `je', `d'r', `ja', `nee' and `neen' are used more by men than by women. Therefore, we decided to count the number of occurrences of these words.
\par
\textit{words used more by females:} Similarly, some other concrete words are used significantly more by women: `ik', `hij', `dadelijk', `vriendelijk', `lelijk', `vrolijk', `eindelijk' and `verschrikkelijk'.
\par
\textit{diminutives count:} We also counted the diminutives appearing in every data example as diminutives are used more by women than by men \cite{keune2012explaining}.

We started by using the classifier with the strongest baseline on unigrams for the 3 out-of-domain validation datasets. This appeared to be the LogReg classifier with an average accuracy of $53.11\%$. We first evaluated all features separately, then combinations of two and three features. Using 4 features or more lead to decline in the accuracy. Interestingly, as can be observed in Table \ref{tbl:featuresLog}, by using a single feature, the \textit{cluster} feature outperformed all other features on all three cross-genre datasets. The highest average accuracy for the traditional models ($53.24\%$) was obtained by combining \textit{clusters}, \textit{words used more by men} and \textit{char 3grams}.

\begin{table}[h!]
\centering
\footnotesize\begin{tabular}{c|l|r|r|r|}
\multicolumn{2}{c}{} & \multicolumn{3}{c}{Features}\\\cline{3-5}
\multicolumn{1}{c}{} & & \bf{Clusters}     & \bf{Char 3-grams} & \bf{Unigrams} \\ \cline{2-5}
\multirow{3}{*}{\rotatebox[origin=c]{90}{Valid}}
& \bf{YouTube}	                & \bf{53.55\%}           & 52.84\%           & 53.28\% \\\cline{2-5}
& \bf{Twitter}	                & \bf{53.00\%}           & 52.93\%           & 50.99\%       \\ \cline{2-5}
& \bf{News}	                    & \bf{53.11\%}           & 52.78\%           & 51.20\%       \\ \cline{2-5}
\end{tabular}
\caption{The performance of the best traditional model (LogReg) on the different cross-genre datasets based on the three best single features: clusters, char 3-grams and unigrams}\label{tbl:featuresLog}
\end{table}

\par
\textbf{The Winning Approach}
From the experiments, it resulted that, both for in-genre and cross-genre experiments, the neural models outperformed (with many configurations) the traditional approaches with different feature sets as well as the statistical language models. Therefore, we decided to continue working with the neural models of which the best ones were SpaCy, LSTMa, and SA. We trained 5 instances of these models, each with a different random seed. In addition, for the LSTMa and SA, 5 instances were trained with pretrained frozen word vectors and 5 without. This resulted in a total of 25 models. 

\par
\textit{Ensembling} We opted for a weighted ensemble, where each model contributes its deviation from randomness, i.e. if the validation score of a model is 0.55, it will have a positive weight of 0.05 ($0.55-0.50$), however, if the validation score is 0.40, the weight will be -0.10 ($0.40-0.50$). Afterwards, we computed the output for every model ($1$ or $-1$) and added the prediction multiplied by the weights.

\section{Results}~\label{sec:results}
In the following paragraphs, we present the results of our final weighted ensemble model.

The shared task allowed to submit 2 sets of results for every scenario, allowing thus for maximum of 12 different models. Our first set of results  (\textbf{E-1}) are presented in Table~\ref{tbl:resultsEnsemble1}. These results are all generated by the ensemble models described above. No external data was used here.

\begin{table}[h!]
\centering
\footnotesize\begin{tabular}{|l|r|r|}
\hline
\bf{Test set}       & \bf{in-genre}         & \bf{cross-genre} \\ \hline
\bf{Twitter}	    & 64.75\%               & 57.89\%     \\ \hline 
\bf{Youtube}	    & 62.47\%               & 56.98\%     \\ \hline
\bf{News}           & 66.60\%               & 53.50\%     \\ \hline\hline
% \bf{KB}             & \slash\slash\slash    & 49.46\%     \\ \hline
\bf{AVG}            & 64.61\%                 & 56.12\%     \\ \hline

\end{tabular}
\caption{The accuracies of our first set of results on the different test sets for in-genre and cross-genre gender prediction (\textbf{E-1}).}\label{tbl:resultsEnsemble1}
\end{table}

Our second set of submissions (\textbf{E-2}) were generated by models trained with external data from CSI~\cite{verhoeven2014clips} and TwiSty~\cite{verhoeven2016twisty}. From Table~\ref{tbl:resultsEnsemble}, it results that, the scores for the in-genre data are on average 8\% higher than those for the cross-genre prediction task. With in-genre prediction scores ranging between $63.49$ and $66.10$, the performance on all three domains are relatively similar, YouTube being the hardest genre to predict.   

\begin{table}[h!]
\centering
\footnotesize\begin{tabular}{|l|r|r|}
\hline
\bf{Test set}       & \bf{in-genre}         & \bf{cross-genre} \\ \hline
\bf{Twitter}	    & 65.01\%               & 55.89\%     \\ \hline 
\bf{Youtube}	    & 63.49\%               & 57.10\%     \\ \hline
\bf{News}           & 66.30\%               & 55.80\%     \\ \hline\hline
% \bf{KB}             & \slash\slash\slash    & 49.46\%     \\ \hline
\bf{AVG}            & 64.94\%                 & 56.26\%     \\ \hline
\end{tabular}
\caption{The accuracies of our best performing submission on the different test sets for in-genre and cross-genre gender prediction. (E-2)}\label{tbl:resultsEnsemble}
\end{table}

With respect to the results obtained in the cross-genre setting, we observe low accuracies ($55.80$--$57.10$). However, during our experiments we observed that the models consistently got an accuracy of more than $50\%$ on all validation data, i.e., better than random predictions, indicating that there are speaker-specific language or style features that aid to identify the gender of the speaker/writer. However, a further exploration of the outputs of the models is needed in order to draw more concrete conclusions. 
%Nevertheless, on the secret dataset (KB), our ensemble model did not reach the 50\% benchmark. Interestingly, none of the other models submitted for this shared task did either. 

\section{Conclusions and Future Work}\label{sec:conclusions}

More and more research has focused on author profiling and gender prediction more particularly. The performance of models largely depends on the languages and domains involved. In this work, we explored different methods and models that achieve state-of-the-art results on various Natural Language Processing tasks and applied them on the task of gender prediction for Dutch. The plethora of empirical data we collected confirms the complexity of this task and we can not help but wonder whether we are to expect further improvements on this type of tasks for Dutch.

\section*{Acknowledgements}\label{sec:ack}

This work has been supported by Dublin City University Faculty of Engineering \& Computing under the Daniel O'Hare Research Scholarship scheme and by the ADAPT Centre for Digital Content Technology, funded under the SFI Research Centres Programme (Grant  13/RC/2106) and Theo Hoffenberg, founder \& CEO of Reverso. 

We would also like to thank the organizers of the shared task.

% include your own bib file like this:
%\bibliographystyle{acl}
%\bibliography{acl2018}
\bibliography{acl2018}
\bibliographystyle{acl_natbib}

\end{document}